\documentclass[10pt, a4paper]{article}

\usepackage[T2A, T1]{fontenc}
\usepackage{amsmath}
\usepackage{lrec-coling2024} 

\usepackage{changes}

\title{Ukrainian Visual Word Sense Disambiguation Benchmark}

\name{Yurii Laba, Yaryna Mohytych, Ivanna Rohulia, Halyna Kyryleyza, \\      
\large{\textbf{Hanna Dydyk-Meush, Oles Dobosevych, Rostyslav Hryniv}}}

\address{Ukrainian Catholic University \\
         2A Kozelnytska st., Lviv, Ukraine, 79026  \\
         \{laba, mohytych.hn, rohulia.hn, kyryleyza, hanna\_dydykmeush, dobosevych, rhryniv\}@ucu.edu.ua\\}

\abstract{
This study presents a benchmark for evaluating the Visual Word Sense Disambiguation (Visual-WSD) task in Ukrainian. The main goal of the Visual-WSD task is to identify, with minimal contextual information, the most appropriate representation of a given ambiguous word from a set of ten images. To construct this benchmark, we followed a methodology similar to that proposed by ~\citet{raganato-etal-2023-semeval}, who previously introduced benchmarks for the Visual-WSD task in English, Italian, and Farsi. This approach allows us to incorporate the Ukrainian benchmark into a broader framework for cross-language model performance comparisons. We collected the benchmark data semi-automatically and refined it with input from domain experts. We then assessed eight multilingual and multimodal large language models using this benchmark. All tested models performed worse than the zero-shot CLIP-based baseline model ~\cite{radford2021learning} used by ~\citet{raganato-etal-2023-semeval} for the English Visual-WSD task. Our analysis revealed a significant performance gap in the Visual-WSD task between Ukrainian and English.
 \\ \newline }

\begin{document}

\maketitleabstract

\section{Introduction}

The rise of Large Language Models (LLMs) represents a notable advancement in Natural Language Processing (NLP), catalyzing outstanding progress in text understanding and synthesis. Building upon this milestone, Multimodal Large Language Models (MLLMs) emerged as a pivotal development. MLLMs exhibit remarkable efficacy across diverse domains, including but not limited to image classification, object recognition, and tasks integrating textual and visual inputs.

Despite the distinguished milestones achieved by MLLMs/LLMs, they confront various issues that can detrimentally impact the performance of the models. One such challenge involves problems associated with hallucination generation \cite{huang2023survey}. This phenomenon frequently leads to producing content that deviates from real-world facts or user inputs. It causes significant challenges for the practical usage of these models and evokes concerns on the reliability of LLMs in real-world applications. Furthermore, MLLMs/LLMs demonstrate notably inferior performance when engaged in processing low-resource languages like Ukrainian.

\begin{figure}[!ht]
\begin{center}
\includegraphics[scale=0.3]{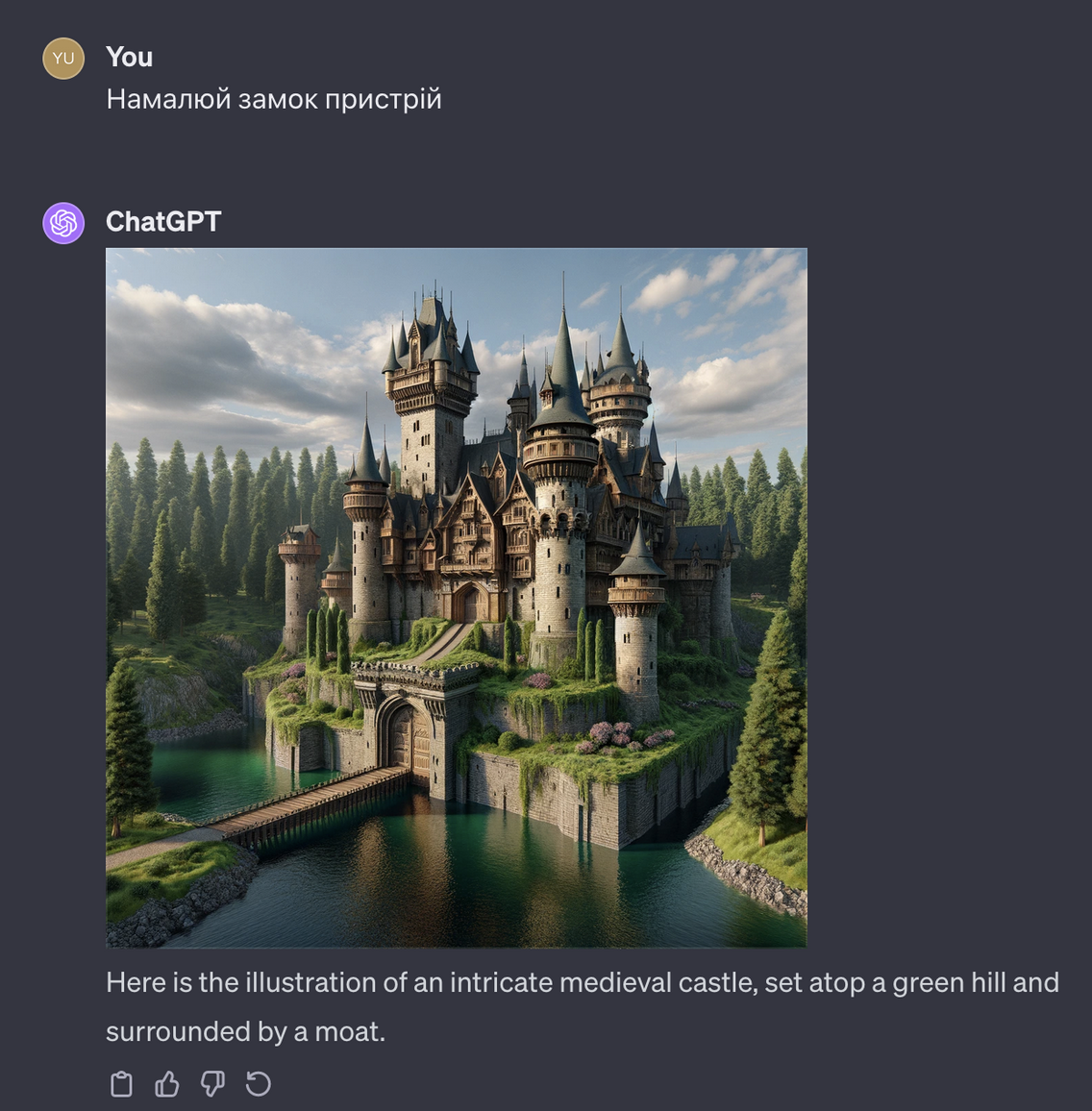} 
\caption{An illustration of GPT4-Vision visual hallucination caused by ambiguous target word.} 
\label{fig.vis_hal}
\end{center}
\end{figure}

In our study, we have opted to examine the extent of hallucinations linked to the utilization of homonyms in the Ukrainian language. Figure~\ref{fig.vis_hal} demonstrates visual hallucination of GPT4-Vision model. This type of hallucination occurred during the generation of an image representing {\fontencoding{T2A}\selectfont Замок} (castle, translit: zamok) with the intended meaning of {\fontencoding{T2A}\selectfont Пристрій} (device, translit: prystriy). In English, this would correspond to the term padlock.

Such hallucinations may arise from several potential reasons. One contributing factor could be the uneven frequency of usage among homonym pairs, wherein certain homonyms are more frequently employed than others. Another contributing factor might be the training of LLMs in multiple languages. The scarcity of available data for low-resource languages often leads subword tokenizers to generate an imbalanced subword vocabulary. As a result, LLMs encounter difficulties in generating high-quality representations for tokens from low-resource languages \cite{hangya-etal-2022-improving, holmstrom-etal-2023-bridging}. Another challenge emerges during domain adaptation. Employing a uniform, single approach to adapt a Language Model (LM) across multiple languages, often referred to as the ``one-size-fits-all'' method, may prove ineffective due to the unique semantic nuances present in each language \cite{grangier-iter-2022-trade}. For instance, a direct translation of a legal term from English to Ukrainian might overlook certain intricate meanings or contextual connotations pertinent to the Ukrainian legal context.


The principal aim of our investigation is to construct a benchmark for gauging the issue of hallucinations related to MLLMs/LLMs concerning homonyms. This is achieved by assessing the efficacy of relevant models in tackling the Visual-WSD task. This task involves providing a target ambiguous word and a restricted context, accompanied by ten images. The Visual-WSD task requires the identification of the most relevant image corresponding to the intended meaning of the ambiguous word. In addition, we give an exhaustive account of a comparative analysis of the performance of several relevant Multilingual MLLMs on the developed benchmark and demonstrate that there is a notable disparity in performance in the Visual-WSD task between Ukrainian and English.


\section{Related Works}

Word Sense Disambiguation (WSD) is a general task of identifying the intended sense of a poly-semantic word in a particular context, typically from a predetermined sense inventory. Although recent advances of LLM's have naturally led to multimodal settings of that task, until now, the research on WSD in the Ukrainian language has been primarily focused on textual modality alone.

One of the approaches \cite{laba-etal-2023-contextual} employs a fine-tuning of LM in a semi-supervised manner to improve its performance on the WSD task in Ukrainian. In their study, the authors indicated that language model-based solutions outperform traditional methods \cite{barba2021consec, moro2014entity}. They also compiled a benchmark based on the Dictionary of Ukrainian Language (SUM) \citeplanguageresource{ukrainian_dictionary}, which can be used to assess the performance of various models on the textual WSD task in Ukrainian.

The popularity of the visual generative models like DALL-E \cite{pmlr-v139-ramesh21a} or Stable Diffusion \cite{rombach2022high} and the abundance of visual information around was probably the reason why Task~1 of the SemEval-2023 conference was on the Visual-WSD. The methodology of the competition (i.e., the dataset and evaluation metrics) is described in~\citet{raganato-etal-2023-semeval}. The datasets are available in English, Italian, and Farsi, enabling testing different approaches to the solution of the Visual-WSD task. 

Most of the suggested solutions were based upon the foundation of the CLIP model \citep{radford2021learning} utilizing Teacher Learning technique \cite{{hinton2015distilling}}. Teacher Learning is a domain-agnostic machine learning technique that transfers knowledge from a pre-trained teacher model to a new student model. This method has been successfully applied not only in multimodal settings but also in various other tasks \cite{reimers-gurevych-2019-sentence, wu-etal-2020-single}.

\citet{carlsson-etal-2022-cross} give a successful example of applying the Teacher Learning technique with Multilingual CLIP. Their approach relies solely on machine translation and thus eliminates the need for visual data in the target language. The proposed objective is to reduce the Mean Squared Error (MSE) between the embeddings produced by the teacher model and the student model for translated texts. Rather than optimizing directly for cosine similarity, as in the original CLIP training, MSE is employed due to its proven effectiveness in providing a better learning response.

A comparable but slightly different approach was proposed by \citet{reimers-gurevych-2020-making}. Their aim was to minimize the MSE between the embeddings generated by the teacher model and those produced by the student model for both the source sentences and their translations.

The recent LLaVA-1.5 model \citep{liu2023improved} utilizes a completely different approach based on  visual instruction tuning. This model operates primarily as a standard causal LM, taking language instructions (a user text prompt) as input and generating a language response. Its ability to process images is facilitated by an independent vision encoder model, which converts images into language tokens that are seamlessly integrated into the user text prompt. The LM and vision encoder of LLaVA are built upon two reference models known as Vicuna \cite{zheng2024judging} and CLIP \cite{radford2021learning}, respectively.

Even though some of the mentioned approaches to the Visual-WSD task are language-agnostic, the efficiency of the respective models in completing instructions in one language cannot be evaluated on datasets in another language, since most challenges are caused by word polysemy, which is typically language-specific. To the best of our knowledge, there are currently no evaluation resources in languages other than those proposed in the SemEval-2023 Task 1, and this hampers research of multilingual and multimodal LMs. That was one of the main motivations for us to create an evaluation dataset for the Visual-WSD task in Ukrainian following the methodology of~\citet{raganato-etal-2023-semeval} and to benchmark on it the available approaches. We hope these resources will facilitate future research in multimodal language models.


\section{Approach}

In this section, we provide a rationale for selecting specific textual and visual data sources and describe the data collection process and semi-automation of the annotation process employed to create the benchmark.

\subsection{Data sources}

Effective evaluation of the Visual-WSD task requires image-word pairs with challenging word instances, e.g. those with multiple meanings (polysemantic words). Ukrainian Wikipedia\footnote{\url{https://en.wikipedia.org/wiki/Ukrainian_Wikipedia}} seemed to be a good source for identifying such words; however, our subsequent analysis revealed multiple problems and shortcomings (misleading links, missing images/sections, irrelevant articles, etc.) in data collected that way. Consequently, we opted to use homonyms listed in reliable dictionary sources.

The dictionary of homonyms of the Ukrainian language \citeplanguageresource{dictionary_of_homonyms} is seemingly the only thorough and reliable research work. The dictionary is only available as a published physical book; with the authors' and publisher's permission, we run the Optical Character Recognition (OCR) software to transform the textual information into a soft copy.

After refining the dictionary post-OCR quality, experts in the Ukrainian language and specialized knowledge in Ukrainian philology performed a thorough selection of homonyms. The selected homonyms are nouns (to optimize the search for visual complementary material), of high usage frequency in the modern Ukrainian language (according to the \citetlanguageresource{grack}), and are full (with the aligned paradigm of forms). 

In the annotation stage, links to the corresponding Wikipedia sources were collected, including the word in the proper sense and the accompanying image. It is worth noting that while Wikipedia provided a convenient source and API for automating data collection process, its reliability was occasionally compromised. There were eminent challenges such as the absence of a direct article for certain words or missing images on the page of word definition. Also, unlike its English counterpart, the Ukrainian Wikipedia often suffers from incomplete information and numerous missing sections.

\subsection{The methodology for constructing the benchmark}

\begin{figure}[!ht]
\begin{center}
\includegraphics[scale=0.4]{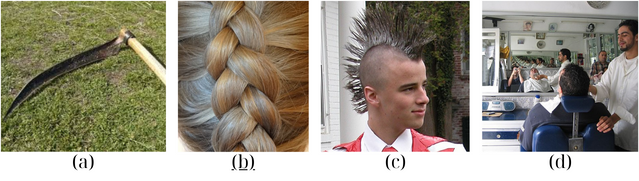} 
\caption{Example of the benchmark entry. The word {\fontencoding{T2A}\selectfont Коса} (en: braid, transl: kosa) is ambiguous. It corresponds to the meaning {\fontencoding{T2A}\selectfont Заплетене волосся; довге волосся} (en: braided hair; long hair, transl: zapletene volossya; dovhe volossya). The word {\fontencoding{T2A}\selectfont Волосся} (en: hair, transl: volossya) is the trigger word. The image that corresponds to the intended meaning is b (underlined). The other three images are examples of negative samples. Note: While the task involves nine negative images, we only display three negative images for simplicity.} 
\label{fig.dataset_sample}
\end{center}
\end{figure}

Each entry in the benchmark includes a target word along with one or multiple trigger words paired with ten unique images: one image corresponds to the intended meaning of the ambiguous target word, serving as a positive sample. The remaining images are negative samples and correspond to
\begin{itemize}
\item alternative interpretations of the ambiguous target word (3 images per entry);
\item similar words within the domain (3 images per entry);
\item randomly selected concepts (3 images per entry).
\end{itemize}

Figure~\ref{fig.dataset_sample} provides a simplified overview of single entry of the benchmark.

We generated positive samples by extracting the title picture of the ambiguous word from its corresponding Wikipedia article. In cases where the article lacked a valid image or any image altogether, domain experts supplied one. Negative samples for each word sense were constructed using other Wikipedia articles from the same domain as the target sense. By employing such a method, we aimed to discover articles about other senses of the ambiguous word, similar concepts within the same domain, and consequently, obtain corresponding images. We also hypothesized that this approach could lead us to discover images from completely different, unrelated concepts.

Using this methodology, we collected forty negative samples for each word sense. Subsequently, domain experts analyzed these samples and retained only the nine most relevant images for each group of negative samples.

We possessed a list of ambiguous words along with their respective definitions provided by domain experts. To generate trigger words for each entry, we tasked domain experts with supplying several words capable of identifying the intended meaning of the word sense when considering the correlation between definition and image. These trigger words were deliberately chosen to be sufficiently challenging so as not to reveal the meaning of the image in isolation; the target word typically remained necessary to comprehend the complete context. This process aimed to ensure a demanding text disambiguation task.

At the time of evaluation, the benchmark included 87 homonyms. However, we are in the process of expanding the homonym list and will update the benchmark accordingly in the future.


\section{Evaluation}

In this section, we explore the metrics utilized to assess model performance on our benchmark and present the results of models evaluation.

To generate predictions, we compare the model embeddings representing the query phrase and those representing each candidate image. The candidate image showing the highest cosine similarity to the query is then identified as the prediction.

In instances where the direct retrieval of embeddings is impossible (e.g., in GPT4-Vision), we prompt the model with both the query and all image candidates, and instruct it to rank the images from the most closely associated with the query to least associated.

\subsection{Evaluation metrics}

To evaluate models' performance, we have used the 
Mean Reciprocal Rank (MRR) and HIT@1 metrics.

Given \( r = [r_1, \ldots, r_n] \) as the image ranking predictions provided by a model, MRR is defined as:
\begin{equation}
\text{MRR} = \frac{1}{n} \sum_{i=1}^{n} \frac{1}{r_i} \times 100\%,
\end{equation}
where $n$ is the number of queries and $r_i$ is the rank of the correct result for the $i$-th query.
Another metric we use 
is the HIT@1 score defined as
\begin{equation}
\text{HIT@1} = \frac{1}{n} \sum_{i=1}^{n} \text{correct}(r_i) \times 100\%,
\end{equation}
where $correct(r_i)$ is an indicator function that equals 1 if $r_i=1$ (i.e., the correct result is ranked first) and 0 otherwise. The HIT@1 metric can also be interpreted as the accuracy of the ranking model.

MRR evaluates the model's efficiency in retrieving relevant images by considering the position of the first relevant image in the ranked list, providing a comprehensive measure of retrieval effectiveness. 

HIT@1 directly measures the model's accuracy in identifying the most relevant image by assessing the proportion of queries for which the top-ranked image is relevant.

\subsection{Results}

\begin{table}
\centering
\begin{tabular}{lll}
\hline
\textbf{Model} & \textbf{HIT@1} & \textbf{MRR}\\
\hline 
\href{https://huggingface.co/M-CLIP/XLM-Roberta-Large-Vit-B-16Plus}{XLM-Roberta-Large-} && \\
\href{https://huggingface.co/M-CLIP/XLM-Roberta-Large-Vit-B-16Plus}{Vit-B-16Plus} & 42.78 & 60.30 \\[2pt]

\href{https://platform.openai.com/docs/models/gpt-4o}{GPT-4o} & 42.52 & 43.75 \\[2pt]

\href{https://huggingface.co/M-CLIP/XLM-Roberta-Large-Vit-L-14}{XLM-Roberta-Large-Vit-L-14} & 40.21 & 58.65 \\[2pt]

\href{https://huggingface.co/M-CLIP/XLM-Roberta-Large-Vit-B-32}{XLM-Roberta-Large-Vit-B-32} & 39.69 & 57.69 \\[2pt]

\href{https://platform.openai.com/docs/guides/vision}{GPT4-Vision} & 38.50 & 45.29 \\[3pt]

\href{https://huggingface.co/M-CLIP/LABSE-Vit-L-14}{LABSE-Vit-L-14} & 35.57 & 54.37 \\[2pt]

\href{https://huggingface.co/sentence-transformers/clip-ViT-B-32-multilingual-v1} {clip-ViT-B-32-multilingual-v1} & 32.99 & 52.46 \\[2pt]

\href{https://cloud.google.com/vertex-ai/generative-ai/docs/model-reference/multimodal-embeddings} {GCP  Multimodal} && \\
\href{https://cloud.google.com/vertex-ai/generative-ai/docs/embeddings/get-multimodal-embeddings#python} {Embeddings} & 22.68 & 41.74 \\[2pt]

\href{https://huggingface.co/llava-hf/llava-1.5-7b-hf} {LLaVA-1.5} & 14.43 & 33.03 \\

\hline \\[-7pt]
\href{https://huggingface.co/sentence-transformers/clip-ViT-B-32-multilingual-v1}{clip-ViT-B-32-multilingual-v1} && \\
(baseline on English language) & 60.48 & 73.88 \\[3pt]
\hline
\end{tabular}
\caption{\label{evaluation_results} The HIT@1 and MRR metrics for multiple multimodal models evaluated on the assembled benchmark and sorted by HIT@1. Baseline results for Visual-WSD in the English language are also included for comparison \cite{raganato-etal-2023-semeval}.}
\end{table}

Table~\ref{evaluation_results} gives an overview of the performance evaluation metrics of multiple multilingual models on the compiled benchmark. 

The results demonstrate that all evaluated models performed less effectively in Ukrainian compared to the English baseline model which highlights a disparity in performance between Ukrainian and English in the Visual-WSD task. 

\section{Conclusion}
This research introduces a benchmark for the Visual-WSD task in the Ukrainian language\footnote{\href{https://drive.google.com/drive/u/1/folders/1hsyOSDTDVHDLhyN-rjEcb4eEAEbGngv2}{U-VWSD benchmark}}. Unlike traditional single-modality benchmarks, we propose an approach that integrates textual and visual modalities into a single benchmark. 


Furthermore, we assessed various suitable multilingual models using the compiled benchmark. Our findings revealed a notable underperformance in the Visual-WSD task for the Ukrainian language compared to English.

\section{Future plans}
We plan to expand the list of homonyms by introducing such units that the neo-lexicography of the Ukrainian language has not yet recorded. Still, they have become a vital part of modern Ukrainian speech (academic or informal). Examples of such words are {\fontencoding{T2A}\selectfont бот} (en: bot, transl: bot) in meaning {\fontencoding{T2A}\selectfont програмний агент} (en: program agent, transl: prohramnyy ahent), {\fontencoding{T2A}\selectfont град} (en: hail, transl: hrad) in meaning {\fontencoding{T2A}\selectfont бойова машина} (en: combat vehicle, transl: boyova mashyna) and many others.

We intend to publish the benchmark and a compiled set of homonyms and their corresponding definitions as online resources, complete with an API for accessing the materials\footnote{\href{https://ucuapps.github.io/ukrainian-vwsd-benchmark/} {U-VWSD web page}}.

Furthermore, we plan to integrate the compiled benchmark for the Ukrainian language into existing benchmarks for other languages, facilitating the research in multilingual and multimodal LLMs.

\section{Limitations}
Currently, we have constructed a benchmark using a limited number of homonyms. Specifically, we have focused on homonyms, which are nouns with a high frequency of usage in Ukrainian. There exist homonyms in the Ukrainian language that remain undocumented in its neo-lexicography. At present, we have excluded these homonyms from our benchmark. Nonetheless, these homonyms constitute a significant component of the Ukrainian language. Therefore, it is highly pertinent to include them in model evaluations, considering their widespread usage by speakers.

\section{Ethical Statement}
The domain experts engaged in our research are proven professionals in Ukrainian philology, ensuring a high standard of work in selecting suitable images and contextual information. The images do not contain any harmful or detrimental content.


\nocite{*}
\section{Bibliographical References}
\label{sec:reference}

\bibliographystyle{lrec-coling2024-natbib}
\bibliography{u-vwsd-bibliography}

\section{Language Resource References}
\label{lr:ref}
\bibliographystylelanguageresource{lrec-coling2024-natbib}
\bibliographylanguageresource{u-vwsd-languageresource}

\end{document}